\documentclass{article}

\usepackage{microtype}
\usepackage{graphicx}
\usepackage{subfigure}
\usepackage{caption}
\usepackage{booktabs} % for professional tables

\usepackage{hyperref}

\usepackage[accepted]{icml2023}

\usepackage{amsmath}
\usepackage{amssymb}
\usepackage{mathtools}
\usepackage{amsthm}
\usepackage{array}
\usepackage{dashrule}
\usepackage{bm}
\usepackage{multirow}
\usepackage{bm}
\usepackage{makecell}
\usepackage{colortbl}
\usepackage{xcolor}
\usepackage[capitalize,noabbrev]{cleveref}

\theoremstyle{plain}

\theoremstyle{definition}

\theoremstyle{remark}

\usepackage[textsize=tiny]{todonotes}

\icmltitlerunning{Evolving Semantic Prototype Improves Generative Zero-Shot Learning}

\begin{document}

\twocolumn[
\icmltitle{Evolving Semantic Prototype Improves Generative Zero-Shot Learning}

% It is OKAY to include author information, even for blind
% submissions: the style file will automatically remove it for you
% unless you've provided the [accepted] option to the icml2023
% package.

% List of affiliations: The first argument should be a (short)
% identifier you will use later to specify author affiliations
% Academic affiliations should list Department, University, City, Region, Country
% Industry affiliations should list Company, City, Region, Country

% You can specify symbols, otherwise they are numbered in order.
% Ideally, you should not use this facility. Affiliations will be numbered
% in order of appearance and this is the preferred way.

%\icmlsetsymbol{equal}{*}

\begin{icmlauthorlist}
\icmlauthor{Shiming Chen}{CMU,MBZUAI}
\icmlauthor{Wenjin Hou}{HUST}
\icmlauthor{Ziming Hong}{USYD}
\icmlauthor{Xiaohan Ding}{Ten}
\icmlauthor{Yibing Song}{Fudan}\\
\icmlauthor{Xinge You}{HUST}
\icmlauthor{Tongliang Liu}{MBZUAI,USYD}
\icmlauthor{Kun Zhang}{CMU,MBZUAI}

%\icmlauthor{}{sch}
%\icmlauthor{}{sch}
\end{icmlauthorlist}

\icmlaffiliation{CMU}{Carnegie Mellon University, Pittsburgh PA, USA}
\icmlaffiliation{MBZUAI}{Mohamed bin Zayed University of Artificial Intelligence, Abu Dhabi,UAE}
\icmlaffiliation{HUST}{Huazhong University of Science and Technology, Wuhan, China}
\icmlaffiliation{Ten}{Tencent AI LAB, Shenzhen, China}
\icmlaffiliation{Fudan}{AI$^3$ Institute, Fudan University, China}
\icmlaffiliation{USYD}{University of Sydney, Sydey, Australia}

\icmlcorrespondingauthor{Xinge You}{youxg@hust.edu.cn}
%\icmlcorrespondingauthor{Firstname2 Lastname2}{first2.last2@www.uk}

% You may provide any keywords that you
% find helpful for describing your paper; these are used to populate
% the "keywords" metadata in the PDF but will not be shown in the document
\icmlkeywords{Machine Learning, ICML}

\vskip 0.3in
]

% this must go after the closing bracket ] following \twocolumn[ ...

% This command actually creates the footnote in the first column
% listing the affiliations and the copyright notice.
% The command takes one argument, which is text to display at the start of the footnote.
% The \icmlEqualContribution command is standard text for equal contribution.
% Remove it (just {}) if you do not need this facility.

\printAffiliationsAndNotice{}  % leave blank if no need to mention equal contribution
%\printAffiliationsAndNotice{\icmlEqualContribution} % otherwise use the standard text.

\begin{abstract}

	In zero-shot learning (ZSL), generative methods synthesize class-related sample features based on predefined semantic prototypes. They advance the ZSL performance by synthesizing unseen class sample features for better training the classifier. We observe that each class's predefined semantic prototype (also referred to as semantic embedding or condition) does not accurately match its real semantic prototype. So the synthesized visual sample features do not faithfully represent the real sample features, limiting the classifier training and existing ZSL performance. In this paper, we formulate this mismatch phenomenon as the visual-semantic domain shift problem. We propose a dynamic semantic prototype evolving (DSP) method to align the empirically predefined semantic prototypes and the real prototypes for class-related feature synthesis. The alignment is learned by refining sample features and semantic prototypes in a unified framework and making the synthesized visual sample features approach real sample features.  After alignment, synthesized sample features from unseen classes are closer to the real sample features and benefit DSP to improve existing generative ZSL methods by 8.5\%, 8.0\%, and 9.7\% on the standard CUB, SUN AWA2 datasets, the significant performance improvement indicates that evolving semantic prototype explores a virgin field in ZSL.
	
	%Generative models (\textit{e.g.}, generative adversarial network (GAN)) have advanced zero-shot learning (ZSL). Studies on the generative ZSL methods typically produce class-specific visual features of unseen classes to mitigate the issue of lacking unseen samples based on
\end{abstract}

\section{Introduction}\label{Sec1}
Zero-shot learning recognizes unseen classes by learning their semantic knowledge, which is transferred from seen classes. Inspired by GAN~\cite{Goodfellow2014GenerativeAN} and VAE~\cite{Kingma2014AutoEncodingVB}, the generative zero-shot learning methods \cite{Arora2018GeneralizedZL} synthesize image sample features related to unseen classes based on the predefined class semantic prototypes (also referred to as semantic embedding in~\cite{Yan2021ZeroNASDG} and semantic condition in~\cite{Xian2019FVAEGAND2AF}). The CNN features are typically utilized to synthesize unseen samples in generative ZSL. The synthesized sample features of unseen classes via generative ZSL benefit the classifier training process and improve the ZSL performance.

%Zero-shot learning (ZSL) recognizes the unseen classes, by transferring semantic knowledge from some known classes to unknown ones. Recently, generative models (\textit{e.g.}, generative adversarial networks (GANs) \cite{Goodfellow2014GenerativeAN} and variational autoencoder (VAE) \cite{Kingma2014AutoEncodingVB}) have been successfully applied in ZSL and achieved promising performance. They synthesize the class-specific images or visual features of unseen classes to mitigate the lack of unseen samples based on the condition of the predefined class semantic prototypes \cite{Arora2018GeneralizedZL
	
\begin{figure*}[t]
		\begin{center}			
			\includegraphics[width=1.0\linewidth]{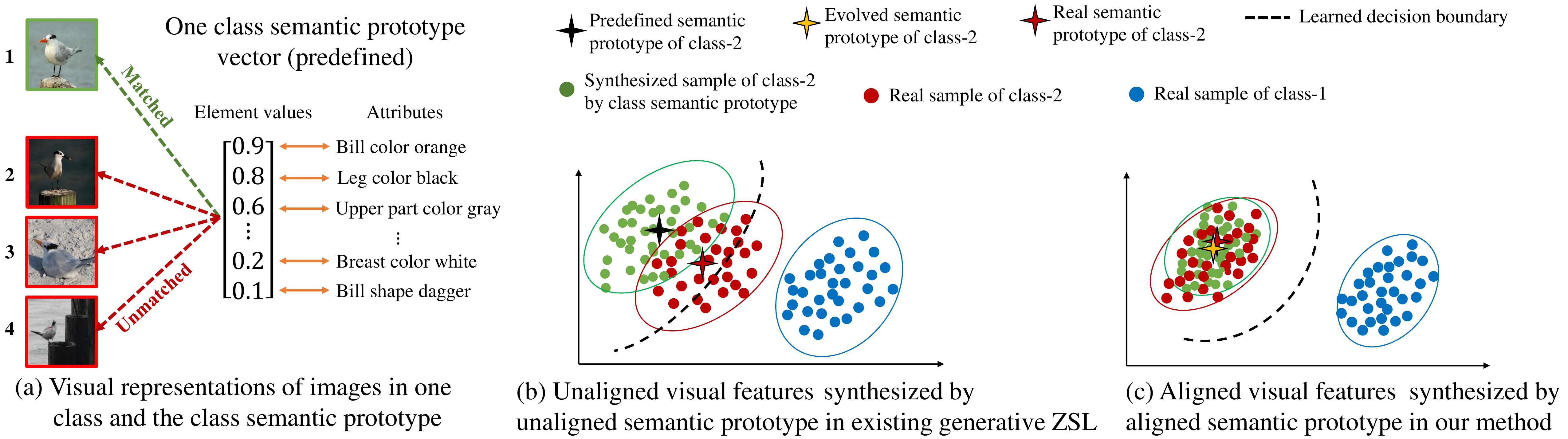}\\            
			\caption{Our motivation. For a specific class, generative ZSL methods synthesize sample features based on the predefined class semantic prototype vector. The misalignment between this prototype and the real one makes this prototype not always match the features of real visual samples, as shown in (a). The visual sample features synthesized by existing generative ZSL methods via this prototype do not faithfully represent the real features, as shown in (b). Our method evolves the predefined prototype by aligning it to the real one, improving the synthesized visual features close to the real visual features for learning a more discriminative classifier, as shown in (c).}
		\label{fig:motivation}
	\end{center}
\end{figure*}

In generative ZSL, the unseen sample features for each class are synthesized by conditioning the class semantic prototype. As the class semantic prototype is predefined empirically, it is not able to accurately represent the real semantic prototype that is not available in practice. Fig.~\ref{fig:motivation}(a) shows four examples. The `bill color orange' attribute in sample 2 does not accurately reflect the bird's bill color, which is mostly dark. Moreover, the attribute `leg color black' in sample 3 does not accurately reflect the bird's leg, which is occluded by its body. A similar phenomenon appears in sample 4, where the attribute `bill color orange' does not effectively describe the bird's bill that is mostly occluded as well. As such, the attributes with their element values defined in the semantic prototype do not properly represent image samples due to color variations or partial occlusions. For an unseen class, its semantic prototype and sample features differ from the real prototype and real sample features, respectively, as shown in Fig.~\ref{fig:motivation}(b). The feature discrepancy limits the classifier training and the generative ZSL performance. We formulate this phenomenon as a Visual-Semantic domain shift problem. To mitigate domain shift, we start by updating the semantic prototype and finally approaching the real one. Thus for an unseen class, sample features from the conditional generator will resemble the real features to benefit classifier training and improve the generative ZSL performance.

%The class-specific unseen sample generations are conditioned on the semantic prototypes. However, there is a limitation in that the empirically predefined semantic prototypes cannot faithfully represent the actual semantic prototypes of visual features (\textit{i.e.}, visual prototypes). As shown in Fig. \ref{fig:motivation}, \textit{i)} the predefined class semantic prototypes are annotated by human, inevitably resulting in some inaccurate annotations (\textit{e.g.}, the attribute “bill color orange” in Sample-2 in Fig. \ref{fig:motivation}(a)), and \textit{ii)} the visual images are with multi-views, resulting in the phenomenon that some importantly annotated attributes do not appear in the visual representations (\textit{e.g.}, the attribute “bill color orange” in Sample-3 and attribute “leg color black” in Sample-4 in Fig. \ref{fig:motivation}(a)). That is, some visual representations of one class are incorrectly mapped with their common predefined semantic prototype. As such, the visual features synthesized by existing generative ZSL methods are far away from their corresponding real visual features and visual prototypes (as shown in Fig. \ref{fig:motivation}(b)), which heavily limits their classification performance. We formulate this phenomenon as a problem of \textit{Visual-Semantic Domain Shift}. Therefore, it is essential to refine the empirical defined semantic prototype. Using an accurate prototype benefits the generator supervision to improve the generative ZSL performance. 

In this paper, we propose evolving the predefined semantic prototypes and aligning them with real ones. The alignment is fulfilled in our dynamic semantic prototype evolving method that jointly refines prototypes and sample features. Through joint refinement, the conditional generator will synthesize visual sample features close to real ones for each class, as shown in Fig.~\ref{fig:motivation}(c). Specifically, our DSP consists of a visual$\rightarrow$semantic mapping network (V2SM) and a visual-oriented semantic prototype evolving network (VOPE). The V2SM network takes sample features as input and outputs the class semantic prototypes. The VOPE network takes the semantic prototype at the current step as input and outputs the semantic prototype at the next step. During training, the generator produces sample features based on the semantic prototype at the current step. These features will be sent to the V2SM network to produce semantic prototypes. These prototypes will supervise VOPE output for prototype evolvement. The updated prototypes at each step will be sent back to the generator as a condition for visual sample feature synthesis. During inference, we concatenate the evolved semantic prototype vectors with visual sample features for semantic enhancement for classification. The visual features will be conditioned by the semantic prototype and mitigate the visual-semantic domain shift problem. The synthesized sample features resemble the real features based on the evolved prototype guidance and improve the classification performance.

We evaluate our DSP method on standard benchmark datasets, including CUB~\cite{Welinder2010CaltechUCSDB2}, SUN~\cite{Patterson2012SUNAD} and AWA2~\cite{Xian2019ZeroShotLC}. The evaluation is conducted by measuring the performance improvement of state-of-the-art ZSL methods (i.e., CLSWGAN~\cite{Xian2018FeatureGN}, f-VAEGAN~\cite{Xian2019FVAEGAND2AF}, TF-VAEGAN~\cite{Narayan2020LatentEF} and FREE~\cite{Chen2021FREE}) upon leveraging our DSP during training and inference. Experiments show that the average improvements of the harmonic mean over these methods are $8.5\%$, $8.0\%$, and $9.7\%$ on CUB, SUN, and AWA2, respectively. The significant performance improvement, as well as the flexibility that DSP can be integrated into most generative ZSL methods, shows that evolving semantic prototype explores one promising virgin field in generative ZSL.

\section{Related Works}\label{Sec2}

{\flushleft \bf Embedding-based Zero-Shot Learning.}
Embedding-based ZSL methods were popular early, they learn a visual$\rightarrow$semantic projection on seen classes that is further transferred to unseen classes \cite{Akata2016LabelEmbeddingFI,Akata2015EvaluationOO,Huynh2020FineGrainedGZ,Chen2021TransZero,Huynh2020CompositionalZL}, Considering the cross-dataset bias between ImageNet and ZSL benchmarks \cite{Chen2021FREE}, embedding-based methods were recently proposed to focus on learning the region-based visual features to enhance the holistic visual features\footnote{Holistic visual features directly extract from a CNN backbone (\textit{e.g.}, ResNet101 \cite{He2016DeepRL}) pretrained on ImageNet.} using attention mechanism \cite{Yu2018StackedSA,Zhu2019SemanticGuidedML,Xie2019AttentiveRE,Huynh2020FineGrainedGZ,Xu2020AttributePN,Chen2022GNDANGN,Chen2021TransZeroCA}. However, these methods depends on the attribute features for attribute localization. As such, \cite{Wang2021DualPP} introduces DPPN to iteratively enhance visual features using the category/visual prototype as supervision. Differently, we jointly and mutually refine the visual features and semantic prototype, enabling significant visual-semantic interactions in generative ZSL.

\begin{figure*}[t]
	\begin{center}			
		\includegraphics[width=0.8\linewidth]{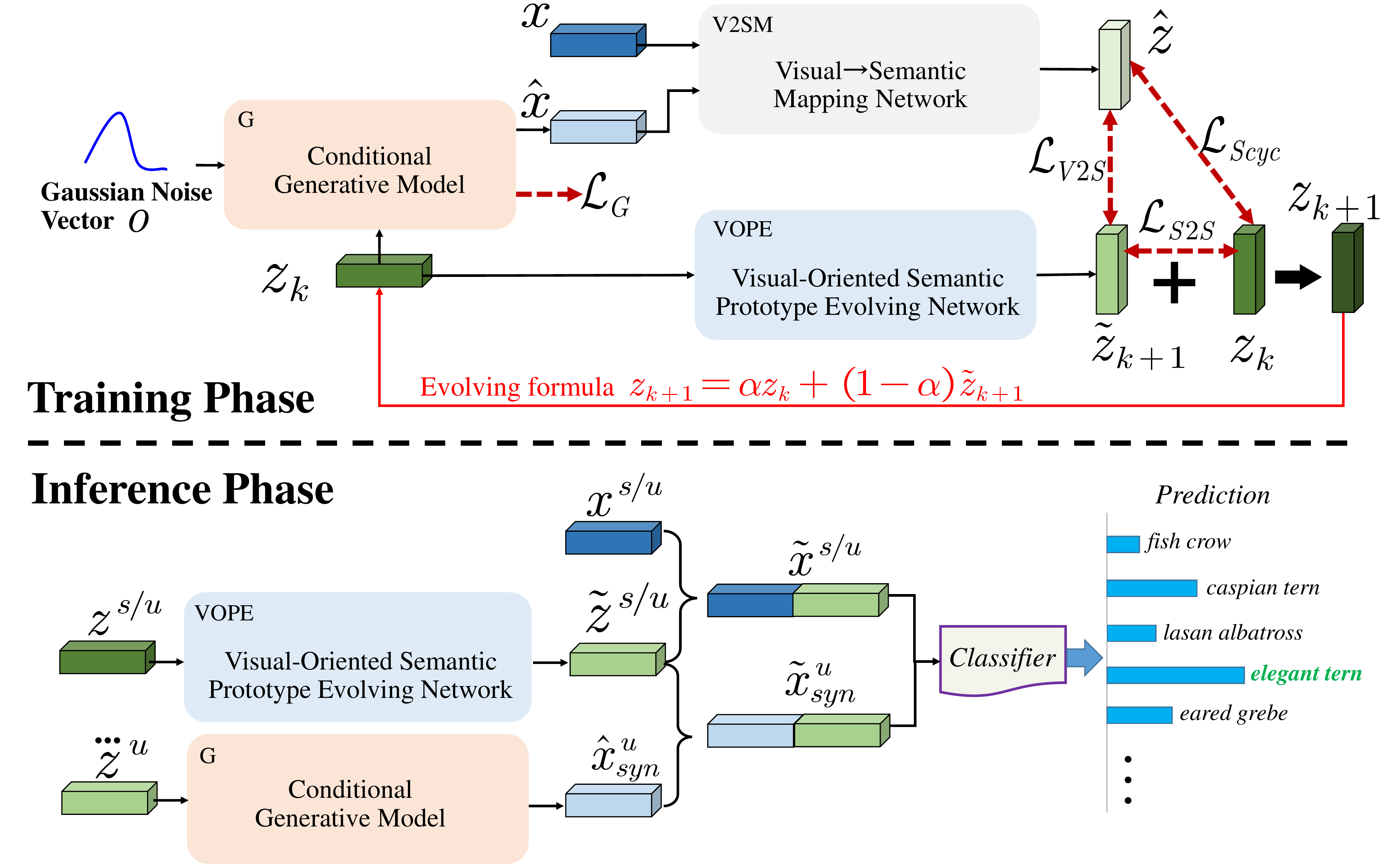}	
		\vspace{-2mm}
		\caption{Dynamic semantic prototype evolvement (DSP). During training, we use V2SM and VOPE networks. V2SM maps sample features into semantic prototypes. VOPE maps dynamic semantic prototypes from the $k$-th step to the $(k+1)$-th step for evolvement. Based on the prototype $z_k$ in the $k$-th step, we use the conditional generator to synthesize sample features $\hat{x}$ and map $\hat{x}$ to prototype $\hat{z}$ via V2SM. We use $\hat{z}$ to supervise VOPE output $\hat{z}_{k+1}$ during evolvement, which brings semantics from sample features. The V2SM and VOPE are jointly trained with the generative model. During inference, we use VOPE to map one input prototype $z^{s/u}$ to its evolved form $\tilde{z}^{s/u}$ for both seen and unseen classes. Besides, we take $\dddot{z}^u= \alpha\cdot z^u+(1-\alpha)\cdot\tilde{z}^u$ as the generator input to synthesize sample features $\hat{x}_{syn}^u$ for unseen classes. Then, we concatenate sample features with their $\tilde{z}^{s/u}$ for semantic enhancement during ZSL classification.}
		\label{fig:framework}
	\end{center}
\end{figure*}

{\flushleft \bf Generative Zero-Shot Learning.}
Since embedding-based methods learn the ZSL classifier only on seen classes, inevitably resulting in the models overfitting seen classes. To tackle this challenge, generative ZSL methods employ the generative models (\textit{e.g.}, VAE and GAN) to generate the unseen visual features for data augmentation \cite{Arora2018GeneralizedZL,Xian2018FeatureGN,Shen2020InvertibleZR,Hong2022SemanticCE}, ZSL is then converted to a supervised classification task further. As such, the generative ZSL methods have shown significant performance and have become very popular recently. However, there are lots of challenges that need to be tackled in generative ZSL. To improve the optimization process of ZSL methods, \cite{Skorokhodov2021ClassNF} introduces class normalization for the ZSL task. \cite{Cetin2022ClosedformSP} introduces closed-form sample probing for generative ZSL. \cite{Chou2021AdaptiveAG} designs a generative scheme to simultaneously generate virtual class labels and their visual features. To enhance the visual feature in generative ZSL methods, \cite{Felix2018MultimodalCG}, \cite{Narayan2020LatentEF} and \cite{Chen2021FREE} refine the holistic visual features by learning a visual$\rightarrow$semantic projection, which encourages conditional generator to synthesize visual features with more semantic information. Orthogonal to these methods, we propose a novel dynamic semantic prototype learning method to tackle the visual-semantic domain shift problem and advance generative ZSL further. 

{\flushleft \bf Visual-Semantic Domain Shift and Projection Domain Shift.}
ZSL learns a projection between visual and semantic feature space on seen classes, and then the projection is transferred to the unseen classes for classification. However, data samples of the seen classes (source domain) and unseen classes (target domain) are disjoint, unrelated for some classes, and their distributions can be different, resulting in a large domain shift \cite{Fu2015TransductiveMZ,Wan2019TransductiveZL,Pourpanah2022ARO}. This challenge is the well-known problem of projection domain shift. To tackle this challenge, inductive-based methods incorporate additional constraints or information from the seen classes \cite{Zhang2019TripleVN,Jia2020DeepUE,Wang2021DualPP}. Besides that, several transductive-based methods have been developed to alleviate the projection domain shift problem \cite{Fu2015TransductiveMZ,Xu2017MatrixTW,Wan2019TransductiveZL}.  We should note that the visual-semantic domain shift is different from the projection domain shift. Visual-semantic domain shift denotes that the two domains (visual and semantic domains) share the same categories and only the joint distribution of input and output differs between the training and test stages. In contrast, the projection domain shift can be directly observed in terms of the projection shift, rather than the feature distribution shift \cite{Pourpanah2022ARO}. As shown in Fig.~\ref{fig:motivation}, we show that visual-semantic domain shift is a bottleneck challenge in generative ZSL and is untouched in the literature.

\section{Dynamic Semantic Prototype Evolvement}\label{Sec3}
Fig.~\ref{fig:framework} shows the framework of our DSP during training and inference. We first revisit ZSL setting and illustrate the spirit of prototype evolvement. Then, we show the detailed framework design in the subsections. 

{\flushleft \bf ZSL setting:}
In ZSL, we have $C^s$ seen classes and the total seen class data $\mathcal{D}^{s}=\left\{\left(x_{i}^{s}, y_{i}^{s}\right)\right\}$, where $x_i^s \in \mathcal{X}$ denotes the $i$-th sample feature (in 2048-dim extracted from a CNN backbone, e.g., ResNet101~\cite{He2016DeepRL}), and $y_i^s \in \mathcal{Y}^s$ is its class label. The $\mathcal{D}^{s}$ is split into a training set $\mathcal{D}_{tr}^{s}$ and a testing set $\mathcal{D}_{te}^{s}$ following~\cite{Xian2019ZeroShotLC}. On the other hand, we have $C^u$ unseen classes with $\mathcal{D}_{te}^{u}=\left\{\left(x_{i}^{u}, y_{i}^{u}\right)\right\}$ data, where $x_{i}^{u}\in \mathcal{X}$ are the sample features of unseen classes, and $y_{i}^{u} \in \mathcal{Y}^u$ are its class label for evaluation. The semantic prototypes are represented by vectors. Each vector corresponds to one class. The total class number is $c \in \mathcal{C}^{s} \cup \mathcal{C}^{u}$. So there are $c$ prototype vectors, and each vector is in the $|A|$ dimension where each dimension is a semantic attribute. Formally, the total semantic prototype vectors are denoted as $z^{c}=\left[z^{c}(1), \ldots, z^{c}(A)\right]^{\top} \in \mathbb{R}^{|A|}$. In the conventional ZSL setting (CZSL), we learn a classifier for unseen classes (i.e., $f_{\rm{CZSL}}: \mathcal{X} \rightarrow \mathcal{Y}^{U}$). Differently in generalized ZSL (GZSL), we learn a classifier for both seen and unseen classes (i.e., $f_{\rm{GZSL}}: \mathcal{X} \rightarrow \mathcal{Y}^{U} \cup \mathcal{Y}^{S}$).

%The problem setting of ZSL is formulated as follows. Assume that we have seen class data $\mathcal{D}^{s}=\left\{\left(x_{i}^{s}, y_{i}^{s}\right)\right\}$ with $C^s$ seen classes, where $x_i^s \in \mathcal{X}$ denotes the $i$-th visual feature with 2048-dim extracted from a CNN backbone (\textit{e.g.}, ResNet101 \cite{He2016DeepRL}), and $y_i^s \in \mathcal{Y}^s$ is the corresponding class label. Note that $\mathcal{D}^{s}$ is divided into training set $\mathcal{D}_{tr}^{s}$ and test set $\mathcal{D}_{te}^{s}$ according to \cite{Xian2019ZeroShotLC}. Another set of unseen classes $C^u$ is $\mathcal{D}_{te}^{u}=\left\{\left(x_{i}^{u}, y_{i}^{u}\right)\right\}$, where $x_{i}^{u}\in \mathcal{X}$ are the visual features of unseen classes, and $y_{i}^{u} \in \mathcal{Y}^u$ are the corresponding labels. A set of class semantic vectors (semantic values annotated by humans according to attributes) of the class $c \in \mathcal{C}^{s} \cup \mathcal{C}^{u}$ with $|A|$ attributes, denoted as $z^{c}=\left[z^{c}(1), \ldots, z^{c}(A)\right]^{\top} \in \mathbb{R}^{|A|}$. In the conventional zero-shot learning (CZSL) setting, we aim to learn a classifier for unseen classes, \textit{i.e.}, $f_{CZ S L}: X \rightarrow Y^{U}$. In contrast, we aim to learn a classifier for seen and unseen classes in the generalized zero-shot learning (GZSL) setting, \textit{i.e.}, $f_{G Z S L}: X \rightarrow Y^{U} \cup Y^{S}$.

{\flushleft \bf Prototype evolvement spirit:} In the training phase of DSP as shown in Fig.~\ref{fig:framework}, we conduct prototype evolvement in VOPE that is supervised by V2SM. In the $k$-th step of evolvement, generator G synthesizes sample features $\hat{x}$ conditioned on the semantic prototype $z_k$. These features are sent to V2SM to produce a prototype $\hat{z}$, which supervises the VOPE output $\tilde{z}_{k+1}$. The $z_k$ and $\tilde{z}_{k+1}$ are then linearly fused as $z_{k+1}$ to finish the $k$-th step prototype evolvement, where semantic sample features $\hat{x}$ are integrated via V2SM mapping to align with the VOPE output $\tilde{z}_{k+1}$. In sum, a semantic prototype helps the generator to synthesize semantic sample features, and these features are mapped via V2SM to supervise VOPE output for prototype evolvement in VOPE. This evolvement mitigates the gap between sample features and semantic prototypes, and aligns the empirically predefined prototype to the real one that is unavailable in practice.

%\noindent \textbf{Overview:}
%As shown in Fig. \ref{fig:framework}, our dynamic semantic prototype learning method (DSP) consists of a visual$\rightarrow$semantic mapping network (V2SM) and a visual-oriented semantic prototype evolving network (VOPE). Entailed on the generator, V2SM maps visual features into the class semantic space, enabling generator synthesizes visual samples with rich semantic information and conducts visual information into the VOPE. Under the supervision of visual signals in V2SM, VOPE iteratively evolves the predefined class semantic prototypes to be a dynamic semantic prototype, which is fed back to the generative network as conditional supervision information to encourage the generator to conduct accurate semantic$\rightarrow$visual mapping. In the inference phase, DSP concatenates the dynamic semantic prototypes and their corresponding visual features (including the real visual features of seen classes and real/synthesized visual features of unseen classes) to enhance visual features, enabling better classification performance. 

\subsection{Generative ZSL Model}\label{Sec3.0} 
Generative ZSL design is to train a conditional generator $G$: $\mathcal{Z} \times \mathcal{O}\rightarrow \hat{\mathcal{X}}$. This generator takes semantic prototypes and Gaussian noise as inputs to synthesize class-specific sample features $\hat{x}$. We train this generator $G$ together with V2SM and VOPE. Then, we use $G$ to synthesize a large number of sample features for unseen classes based on semantic prototypes $z^u$. These synthesized sample features will be utilized to train a standard classifier (e.g., softmax). This classifier then predicts the testing data from unseen classes for ZSL performance evaluations. 

%In our work, we focus on generative ZSL \cite{Xian2018FeatureGN,Xian2019FVAEGAND2AF}. The main goal of generative ZSL is to learn a conditional generator G: $\mathcal{Z} \times \mathcal{O}\rightarrow \tilde{\mathcal{X}}$, which takes the class semantic vectors/prototypes $z\in \mathcal{Z}$ and Gaussian noise vectors $o\in \mathcal{O}$ as inputs, and synthesizes the class-specific visual feature samples $\tilde{x}\in \tilde{\mathcal{X}}$. Once such a conditional generative model is learned, we can take it to synthesizes a large number of class-specific visual feature samples for unseen classes based on the semantic vector $z^{u}$, and the final classifier over $y\in\mathcal{Y}$ can be obtained using any standard supervised classifier (\textit{e.g.}, softmax). 

\subsection{Visual$\rightarrow$Semantic Mapping Network (V2SM)}\label{Sec3.1} 
We design V2SM to map sample features to the semantic prototype. The sample features with semantics, synthesized by generator $G$ conditioned on semantic prototypes, will be mapped via V2SM for prototype supervision. V2SM is a multilayer perceptron (MLP) with a residual block. The detailed network structure is in Appendix~\ref{appdx-A}. V2SM maps real or synthesized sample features ($x$ or $\hat{x}$ in 2048-dim) to the semantic prototypes ($\hat{z}_{real}$ or $\hat{z}_{syn}$ in $|A|$-dim). The mapping can be written as:
\begin{equation}\label{eq:V2S}
	\hat{z}_{real} = \text{V2SM}(x) \quad\text{or}\quad \hat{z}_{syn} = \text{V2SM}(\hat{x})
\end{equation}

where $\hat{z}_{real}\cup\hat{z}_{syn}=\hat{z}$, and $\hat{z}\in \mathbb{R}^{|A|}$. The conditional generator $G$ (i.e., semantic$\rightarrow$visual mapping) and V2SM conduct a semantic cycle network, i.e., $z\rightarrow G(o,z)=\hat{x}$ and $\hat{x}\rightarrow\text{V2SM}(\hat{x})=\hat{z}$, where $o$ is a random Gaussian noise vector with $|A|$-dim. As such, V2SM encourages $G$ to synthesize sample features with semantics. We propose the semantic cycle-consistency loss to improve V2SM mapping, which can be written as:
\begin{equation}\label{eq:L_Scyc}
	\mathcal{L}_{Scyc}= \mathbb{E}\left[\|\hat{z}_{real}-z_{k}\|_{1}\right]+\mathbb{E}\left[\|\hat{z}_{syn}-z_{k}\|_{1}\right],
\end{equation}
where $\mathbb{E}(\cdot)$ is the average value computed by using all the training samples, this operator is also utilized in the following equations, $z_{k}$ is the dynamic semantic prototype from VOPE at the $k$-th step, which will be illustrated in Sec.~\ref{Sec3.2}. This loss is different from TF-VAEGAN~\cite{Narayan2020LatentEF} and FREE~\cite{Chen2021FREE} where the predefined semantic prototypes are utilized as semantic supervision. Since the dynamic semantic prototype is closer to the real prototype, it supervises V2SM and $G$ more accurately and thus further improves TF-VAEGAN and FREE.  

% We find that $z$, $\hat{z}$ and $z_k$ are closely affected by each other. $\hat{z}=\text{V2SM}(x)$ is mapped from the visual features in V2SM and used to convey the visual signal to supervise VOPE, enabling $z$ to be progressively evolved as dynamic semantic prototype $z_k=\alpha z + (1-\alpha) \text{VOPE}(z_{k-1})$ ($z_k=z$ when k=0). As such,  $z_k$ IS closer to its corresponding visual prototype, and serves as the accurate condition information that encourages the generator to synthesize reliable visual features.

\subsection{Visual-Oriented Semantic Prototype Evolving Network (VOPE)}\label{Sec3.2} 
We propose VOPE to refine semantic prototypes under the supervision of $\hat{z}$. After refinement, the dynamic semantic prototype is closer to the real prototype, and the visual-semantic domain shift is alleviated as shown in Fig.~\ref{fig:motivation}(c). As the predefined semantic prototypes are empirically annotated, we refine them progressively via $\hat{z}$ that is mapped from sample features with semantics. The structure of VOPE is designed as an MLP network with a residual block. 
This block acts as a routing gate controlled by channel attention. The channel attention evolves the element value of attributes in semantic prototypes.
The detailed network structure is in Appendix~\ref{appdx-B}. VOPE maps the semantic prototype $z_k$ to another semantic prototype $\tilde{z}_{k+1}$ at the $k$-th step of evolvement. The mapping can be written as:
\begin{equation}\label{eq:S2S}
	\tilde{z}_{k+1}= \text{VOPE}(z_k).
\end{equation}
Note that initially (i.e., $k=0$), $z_k$ is the predefined semantic prototype that is empirically annotated. To enable $\tilde{z}_{k+1}$ evolved under the guidance from $\hat{z}$ that is mapped from semantic sample features, we propose a visual$\rightarrow$semantic loss as follows:
\begin{equation}\label{eq:L_V2S}
	\mathcal{L}_{V2S} =\mathbb{E}\left[1- \operatorname{Cosine}(\hat{z},\tilde{z}_{k+1})\right]
\end{equation}
where we use the cosine similarity as a weak constraint to conduct the semantic transfer from the visual feature domain to the prototype domain. Meanwhile, we propose the semantic reconstruction loss as follows:
\begin{equation}\label{eq:L_S2S}
	\mathcal{L}_{S2S} = \mathbb{E}\left[\|\tilde{z}_{k+1}-z_{k}\|_{1}\right]
\end{equation}
where $\tilde{z}_{k+1}$ is the VOPE output at the $(k+1)$-th step. The final dynamic semantic prototype at the $(k+1)$-th can be computed as:
\begin{equation}\label{eq:DSP}
	z_{k+1} = \alpha\cdot z_{k} +(1-\alpha)\cdot \tilde{z}_{k+1}
\end{equation}
where $\alpha$ is a scale value and is set as 0.9 for a smooth moving average update. As a result, $z_k$ is dynamically evolved and fed back to VOPE as input to start the $(k+1)$-th step.

%Considering that the empirically predefined semantic prototype provides rich semantic priori, $\alpha$ is set to relatively large (\textit{i.e.}, $\alpha=0.9$) for preserving its original information. Such a setting i) encourages the dynamic semantic prototype to evolve progressively under the supervision of visual information, and ii) facilitates model optimization. %After training, the dynamic semantic prototypes for all classes can be learned using VOPE, \textit{i.e.}, {\color{blue}$\{z_{k+1}^{1},z_{k+1}^{2},\cdots,z_{k+1}^{C}\}$}. During the inference phase, {\color{blue}$z_{k+1}^{C^u}$} is further used as conditional class semantic supervision for the generator to synthesize visual features for unseen classes.

\subsection{Training and Inference}\label{Sec3.3}

{\flushleft \bf DSP objective loss function.} 
We train V2SM and VOPE jointly with the generator $G$. The total loss function can be written as follows:
\begin{equation}
	\label{eq:Loss_total}
	\mathcal{L}_{total} = \mathcal{L}_G + \lambda_{Scyc}\mathcal{L}_{Scyc} + \lambda_{V2S}\mathcal{L}_{V2S} + \lambda_{S2S}\mathcal{L}_{S2S}
\end{equation}
where $\mathcal{L}_G$ is the loss of generative model $G$, $\lambda_{Scyc}$, $\lambda_{V2S}$, $\lambda_{S2S}$ are the scale values controlling the influence of each loss term. We set $\lambda_{Scyc}$ equal to $\lambda_{S2S}$ as they both weigh semantic reconstruction loss. In our DSP, by using this loss function, we are effective in training various conditional generators $G$ (i.e., CLSWGAN~\cite{Xian2018FeatureGN}, f-VAEGAN~\cite{Xian2019FVAEGAND2AF}, TF-VAEGAN~\cite{Narayan2020LatentEF} and FREE~\cite{Chen2021FREE}). After training $G$, we are effective in synthesizing sample features that are closely related to unseen class prototypes for the classifier training. This classifier then makes well predictions to achieve superior performance on the ZSL benchmarks. In the following, we illustrate the inference phase where our DSP synthesizes features for training the classifier. 

{\flushleft \bf Sample features synthesis for unseen classes.}
After DSP training, we use VOPE to update the predefined semantic prototypes of unseen classes via Eq.~\ref{eq:S2S} and Eq.~\ref{eq:DSP}. The prototype update can be written as follows:
\begin{equation}
	\dddot{z}^u=\alpha z^u+ (1-\alpha)\text{VOPE}(z^u)
\end{equation}
where $\dddot{z}^u$ is the updated semantic prototype of unseen classes. We use $\dddot{z}^u$ as the condition in $G$ to synthesize sample features of unseen classes. The sample feature synthesis can be written as follows:
\begin{equation}\label{eq:synthesize_feature}
	\hat{x}_{syn}^u = G(o,\dddot{z}^u), 
\end{equation}
where $o$ is a random Gaussian noise vector in $|A|$-dim.

{\flushleft \bf Sample features enhancement for all classes.}
We enhance sample features by concatenating them and the semantic prototypes. The enhanced features are more semantically related to the classes and alleviate the cross-dataset bias. For seen classes, we concatenate training data as $\tilde{x}_{tr}^s=[x_{tr}^s, \tilde{z}^s]$, and testing data as $\tilde{x}_{te}^s=[x_{te}^s, \tilde{z}^s]$, respectively For unseen classes, we concatenate training data as $\tilde{x}_{syn}^u = [\hat{x}_{syn}^u, \tilde{z}^u]$, and testing data as $\tilde{x}_{te}^u = [x_{te}^u, \tilde{z}^u]$, respectively. The $x_{tr}$ is the real visual sample feature in the training set $\mathcal{D}_{tr}^{s}$,  $x_{te}$ is the real visual sample feature in the test sets $\mathcal{D}_{te}^{s}$ and $\mathcal{D}_{te}^{u}$. The $\tilde{z}^s \cup \tilde{z}^u = \tilde{z}$. These concatenated features, with semantic prototype conditions, will be utilized for the classifier training and prediction. 

{\flushleft \bf ZSL classifier training and prediction.}	
The enhanced sample features $\tilde{x}_{tr}^s$ and $\tilde{x}_{syn}^u$ are utilized for the ZSL classifier training (e.g., softmax). We denote the classifier as $f$. For conventional ZSL, we can train the classifier as $f_{\rm{CZSL}}: \mathcal{\tilde{X}} \rightarrow \mathcal{Y}^{u}$. For GZSL, we train the classifier as $f_{GZSL}: \mathcal{\tilde{X}} \rightarrow \mathcal{Y}^{s} \cup \mathcal{Y}^{u}$. After classifier training, we use $\tilde{x}_{te}^s$ and $\tilde{x}_{te}^u$ to evaluate the classifier predictions. Note that our DSP is inductive as there are no real visual sample features of unseen class for ZSL classifier training.

%We employ $\tilde{x}_{tr}^s$ and $\tilde{x}_{syn}^u$ to lean a classifier (\textit{e.g.}, softmax), \textit{i.e.}, $f_{c z s l}: \mathcal{\tilde{X}} \rightarrow \mathcal{Y}^{s} \cup \mathcal{Y}^{u}$ or $f_{g z s l}: \mathcal{\tilde{X}} \rightarrow \mathcal{Y}^{s} \cup \mathcal{Y}^{u}$. Once the classifier is trained, we use $\tilde{x}_{te}^s$ and $\tilde{x}_{te}^u$ to test the model further. Note that our DSP is an inductive method as we do not use the real visual features of unseen classes for model optimization.\vspace{-3mm}

\section{Experiments}\label{Sec4}
In this section, we illustrate the experimental configurations, compare our DSP integrations with existing ZSL methods, and conduct ablation studies. 

\subsection{Experimental Configurations}
{\flushleft \bf Benchmark datasets.}
Our experiments are on four standard ZSL datasets including CUB \cite{Welinder2010CaltechUCSDB2}, SUN \cite{Patterson2012SUNAD}, FLO \cite{Nilsback2008AutomatedFC}) and AWA2 \cite{Xian2019ZeroShotLC}. In CUB, there are 11,788 images and 200 bird classes, where 150 are seen and 50 are unseen. The attributes (i.e., prototype dimension) are 312. In SUN, the image number, scene class number, seen/unseen class numbers, and attributes are 14,340, 717, 645/72, and 102, respectively. In FLO, these numbers are 8,189, 102, 82/20, and 1024, respectively. In AWA2, these numbers are 37,322, 50, 40/10, and 85, respectively. These configurations are the same for all the methods.

%We evaluate the effectiveness of our DSP on three well-known ZSL benchmark datasets, \textit{i.e.}, two fine-grained datasets ( CUB \cite{Welinder2010CaltechUCSDB2}, SUN \cite{Patterson2012SUNAD} and FLO \cite{Nilsback2008AutomatedFC}) and one coarse-grained dataset (AWA2 \cite{Xian2019ZeroShotLC}). CUB consists of 11,788 images of 200 bird classes (seen/unseen classes = 150/50) captured by 312 attributes. SUN includes 14,340 images of 717 scene classes (seen/unseen classes = 645/72) described by 102 attributes. AWA2 contains 37,322 images of 50 animal classes (seen/unseen classes = 40/10) captured by 85 attributes. FLO inludes 8,189 images of 102 flower classes (seen/unseen classes = 82/20) captured by 1024 attributes.

\begin{table*}[t]
	\centering  
	\caption{State-of-the-art comparisons on CUB, SUN, AWA2, and FLO under GZSL settings. Embedding-based methods are categorized as $\clubsuit$, and generative methods are categorized as $\spadesuit$. The best and second-best results are marked in \textbf{\color{red}Red} and \textbf{\color{blue}Blue}, respectively.}
	\vspace{2mm}
	\resizebox{\linewidth}{!}{\small
		\begin{tabular}{l|l|c|ccl|ccl|ccl|ccl}
			\hline
			&\multirow{2}{*}{\textbf{Methods}}&Venue
			&\multicolumn{3}{c|}{\textbf{CUB}}&\multicolumn{3}{c|}{\textbf{SUN}}&\multicolumn{3}{c|}{\textbf{AWA2}}&\multicolumn{3}{c}{\textbf{FLO}}\\
			\cline{4-6}\cline{7-9}\cline{10-12}\cline{13-15}
			&&&\rm{U} & \rm{S} & \rm{H} &\rm{U} & \rm{S} & \rm{H} &\rm{U} & \rm{S} & \rm{H} &\rm{U}  & \rm{S}  & \rm{H} \\
			\hline
			\multirow{2}*{
				\begin{tabular}{c}
					\vspace{-2cm}$\clubsuit$
				\end{tabular}
			}
			&SGMA~\cite{Zhu2019SemanticGuidedML}&NeurIPS'19&36.7 &71.3 &48.5&--&--&--&37.6&87.1&52.5&--&--&--\\
			&AREN~\cite{Xie2019AttentiveRE}&CVPR'19&38.9& \textbf{\color{red}78.7}& 52.1&19.0 &38.8 &25.5& 15.6&\textbf{\color{red}92.9}& 26.7&--&--&--\\
			&CRnet~\cite{Zhang2019CoRepresentationNF}&ICML'19& 45.5 &56.8& 50.5&34.1 &36.5 & 35.3& 52.6 & 78.8 & 63.1&--&--&--\\
			&APN~\cite{Xu2020AttributePN}&NeurIPS'20&65.3& 69.3& 67.2& 41.9 &34.0&37.6&56.5& 78.0 &65.5&--&--&--\\
			&DAZLE~\cite{Huynh2020FineGrainedGZ}&CVPR'20& 56.7&59.6&58.1&\textbf{\color{blue}52.3}&24.3&33.2&60.3&75.7&67.1&--&--&--\\
			&CN~\cite{Skorokhodov2021ClassNF}&ICLR'21& 49.9&50.7& 50.3&44.7 &\textbf{\color{red}41.6} &43.1&60.2&77.1 &67.6&--&--&--\\
			&TransZero~\cite{Chen2021TransZero}&AAAI'22&\textbf{\color{red}69.3}& 68.3&\textbf{\color{red} 68.8}& 52.6&33.4& 40.8& 61.3&82.3& 70.2&--&--&--\\
			&MSDN~\cite{Chen2022MSDN}&CVPR'22&\textbf{\color{blue}68.7}& 67.5& \textbf{\color{blue}68.1}& 52.2& 34.2& 41.3& 62.0& 74.5& 67.7&--&--&--\\
			&I2DFormer~\cite{Naeem2022I2D}&NeurIPS'22&35.3&57.6& 43.8&-- & --& --& \textbf{\color{red}66.8}&76.8& 71.5&35.8&\textbf{\color{red}91.9}&51.5\\
			\hline
			\multirow{2}*{
				\begin{tabular}{c}
					\vspace{-3cm}$\spadesuit$
				\end{tabular}
			}
			&f-VAEGAN~\cite{Xian2019FVAEGAND2AF}&CVPR'18&48.7&58.0&52.9&45.1&38.0&41.3&57.6&70.6&63.5&56.8&74.9&64.6\\
			&TF-VAEGAN~\cite{Narayan2020LatentEF}&CVPR'19&53.7&61.9&57.5&48.5&37.2&42.1&58.7&76.1&66.3&62.5&84.1&71.7\\
			&LsrGAN~\cite{Vyas2020LeveragingSA}&ECCV'20&48.1&59.1& 53.0&44.8&37.7& 40.9&54.6&74.6& 63.0&--&--&--\\
			&AGZSL~\cite{Chou2021AdaptiveAG}&ICLR'21&48.3&58.9&53.1&29.9&40.2&34.3&\textbf{\color{blue}65.1}&78.9& 71.3&--&--&--\\
			&FREE~\cite{Chen2021FREE}&ICCV'21&54.9&60.8&57.7&47.4&37.2&41.7&60.4&75.4&67.1&\textbf{\color{red}67.4}&84.5&\textbf{\color{blue}75.0}\\
			&GCM-CF~\cite{Yue2021CounterfactualZA}&CVPR'21&61.0&59.7& 60.3&47.9& 37.8& 42.2&60.4& 75.1& 67.0&--&--&--\\
			&HSVA~\cite{Chen2021HSVA}&NeurIPS'21&52.7&58.3& 55.3& 48.6& 39.0& \textbf{\color{blue}43.3}&  59.3& 76.6& 66.8&--&--&--\\
			&ICCE~\cite{Kong_2022_CVPR}&CVPR'22&67.3& 65.5& 66.4&--&--& --&65.3&82.3& \textbf{\color{blue}72.8}&66.1& 86.5& 74.9\\
			&FREE+ESZSL~\cite{Cetin2022ClosedformSP}&ICLR'22&51.6& 60.4& 55.7&48.2&36.5 &41.5&51.3&78.0& 61.8&65.6&82.2&72.9\\
			&TF-VAEGAN+ESZSL~\cite{Cetin2022ClosedformSP}&ICLR'22&51.1& 63.3& 56.6&44.0&39.7& 41.7&55.2& 74.7& 63.5&63.5& 83.2&72.1\\
			\cline{2-15}
			&f-VAEGAN~\cite{Xian2019FVAEGAND2AF} + \textbf{DSP}&Ours&62.5&\textbf{\color{blue}73.1}&67.4& \textbf{\color{red}57.7}&\textbf{\color{blue}41.3}&\textbf{\color{red}48.1}&63.7&\textbf{\color{blue}88.8}&\textbf{\color{red}74.2}&\textbf{\color{blue}66.2}&\textbf{\color{blue}86.9}&\textbf{\color{red}75.2}\\
			\hline
	\end{tabular} }
	\label{table:SOTA}
\end{table*}

{\flushleft \bf Implementation details.}
We extract image visual features in 2048-dim from the top-layer pooling units of a CNN ResNet-101~\cite{He2016DeepRL} encoder. The encoder is pretrained on the ImageNet dataset. In the ablation study, we use f-VAEGAN as a baseline and validate the effectiveness of our DSP. The training and validation splits are from \cite{Xian2019ZeroShotLC}. We synthesize 150, 800, and 3400 sample features for each unseen class during classifier training for SUN, CUB, and AWA2 datasets, respectively. The $\lambda_{Scyc}$ and $\lambda_{S2S}$ are set with the same value, which are 0.1, 0.01, and 0.001 for CUB, SUN, and AWA2, respectively. The $\lambda_{V2S}$ is 0.6, 0.6, and 1.0 for CUB, SUN, and AWA2, respectively. The settings on other baselines are in Appendix~\ref{appdx-C}.

{\flushleft \bf Evaluation protocols.}
During testing (i.e., ZSL classification), we follow the evaluation protocols in~\cite{Xian2019ZeroShotLC}. In the CZSL setting, we calculate the top-1 accuracy of the unseen class, which is denoted as $\bm{acc}$. In the GZSL setting, we measure the top-1 accuracy on seen and unseen classes, denoted as $\bm{S}$ and $\bm{U}$, respectively. We also compute the harmonic mean of $\bm{S}$ and $\bm{U}$ to evaluate the GZSL performance. The harmonic mean can be computed as $\bm{H} = (2 \times \bm{S}\times \bm{U})/(\bm{S}+\bm{U})$.

%Their harmonic mean (defined as $\bm{H} = (2 \times \bm{S}\times \bm{U})/(\bm{S}+\bm{U})$) are a better protocols in the GZSL setting. As such, we typically take the $\bm{H}$ to discuss the performance of ZSL methods in the GZSL setting.

\subsection{Comparisons with State-of-the-art Methods}
We integrate our DSP on f-VAEGAN, and compare it with state-of-the-art methods. Table \ref{table:SOTA} shows evaluation results. The comparing methods include
non-generative methods (e.g., SGMA~\cite{Zhu2019SemanticGuidedML}, AREN~\cite{Xie2019AttentiveRE}, CRnet~\cite{Zhang2019CoRepresentationNF}, APN~\cite{Xu2020AttributePN}, DAZLE~\cite{Huynh2020FineGrainedGZ}, CN~\cite{Skorokhodov2021ClassNF}, TransZero~\cite{Chen2021TransZero}, MSDN~\cite{Chen2022MSDN}, and I2DFormer~\cite{Naeem2022I2D}) and generative methods (e.g.,  f-VAEGAN~\cite{Xian2019FVAEGAND2AF}, TF-VAEGAN~\cite{Narayan2020LatentEF},  AGZSL~\cite{Chou2021AdaptiveAG}, GCM-CF~\cite{Chou2021AdaptiveAG}, HSVA~\cite{Chen2021HSVA}, ICCE~\cite{Kong_2022_CVPR}, and ESZSL~\cite{Cetin2022ClosedformSP}). Compared to the generative methods, our f-VAEGAN+DSP setting achieves the best results on all datasets, (i.e., SUN ($\bm{H}=48.1$), AWA2 ($\bm{H}=74.2$) and FLO ($\bm{H}=75.2$)). For instance, our f-VAEGAN+DSP outperforms the latest generative method (i.e., TF-VAEGAN+ESZSL \cite{Cetin2022ClosedformSP}) by a large margin, resulting in the improvements of harmonic mean by 10.8\%, 6.4\%, and 10.7\% on CUB, SUN and AWA2, respectively. Moreover, our DSP effectively improves generative ZSL methods to achieve similar performance to the embedding-based methods. These results consistently demonstrate the superiority and great potential of our DSP in generative ZSL. 

\begin{table}
	\centering  		
	\caption{Comparison with generative ZSL methods on the CUB, SUN, and AWA2 datasets under CZSL setting.}
	\vspace{2mm}
	\resizebox{\linewidth}{!}{\small
		\begin{tabular}{l|c|c|c}
			\hline
			\multirow{2}{*}{\textbf{Methods}} 
			&\multicolumn{1}{c|}{\textbf{CUB}}&\multicolumn{1}{c|}{\textbf{SUN}}&\multicolumn{1}{c}{\textbf{AWA2}}\\
			\cline{2-4}
			&\rm{acc} &\rm{acc} & \rm{acc}\\
			\hline
			%DCN~\cite{Liu2018GeneralizedZL}&56.2& 61.8&65.2\\
			CLSWGAN~\cite{Xian2018FeatureGN}& 57.3&60.8& 68.2\\
			f-VAEGAN~\cite{Xian2019FVAEGAND2AF}&61.0&64.7&71.1\\
			CADA-VAE~\cite{Schnfeld2019GeneralizedZA} &59.8&61.7&63.0\\
			Composer~\cite{Narayan2020LatentEF} &\textbf{69.4}&62.6&71.5\\
			FREE~\cite{Chen2021FREE} &64.8&65.0&68.9\\
			HSVA~\cite{Chen2021HSVA}& 62.8& 63.8&70.6\\
			\hline
			f-VAEGAN + DSP & 62.8&\textbf{68.6}& \textbf{71.6}\\
			\hline
	\end{tabular} }
	\label{table:SOTA-CZSL} \vspace{-5mm}
\end{table}

Besides evaluating under the GZSL setting, we also compare our f-VAEGAN+DSP with the generative methods under the conventional ZSL setting (CZSL). Table~\ref{table:SOTA-CZSL} shows the evaluation results. Our f-VAEGAN+DSP achieves the accuracies of 62.8, 68.6, and 71.6 on CUB, SUN, and AWA2, respectively. These results are competitive compared to the generative ZSL methods under the CZSL setting. Table~\ref{table:SOTA-CZSL} indicates that our DSP is effective in mitigating the semantic-visual domain shift problem for generative ZSL methods in the CZSL setting.

\begin{table*}[t]
	\small
	\centering
	\caption{Ablation study on loss terms, smooth evolvement, and feature enhancement of our DSP. The baseline is f-VAEGAN.}
	\label{table:ablation}        
	\vspace{2mm}
	\resizebox{0.85\textwidth}{!}
	{
		\begin{tabular}{l|ccc|ccc}                
			\hline
			\multirow{2}*{Configurations} &\multicolumn{3}{c|}{\textbf{CUB}} &\multicolumn{3}{c}{\textbf{SUN}}\\
			\cline{2-4}\cline{5-7}
			&\rm{U} & \rm{S} & \rm{H} &\rm{U} & \rm{S} & \rm{H} \\
			\hline
			baseline  &48.7&58.0&52.9&45.1&38.0&41.3\\
			%baseline+DSP(V2SM) & 51.0&55.4&	53.0&--&--&--\\
			baseline+DSP (w/o $\mathcal{L}_{Scyc}$)&60.0&63.9&61.9&57.4&38.4&46.0\\
			baseline+DSP (w/o $\mathcal{L}_{S2S}$)&61.9&69.6&65.5&58.1&37.2&45.4\\
			baseline+DSP (w/o $\mathcal{L}_{V2S}$)&58.8&61.0&59.9&55.1&34.8&42.7\\
			baseline+DSP (w/o  smooth evolvement) & 54.4&55.3&54.8&50.3&36.2&42.1\\
			baseline+DSP (w/o  enhancement) & 52.4&53.9&53.1&54.2&35.0&42.5\\
			baseline+DSP (full) &62.5&73.1&67.4&57.7&41.3&48.1\\
			\hline
		\end{tabular}
	}
\end{table*}

\begin{table*}
	\centering  
	\caption{Evaluation of DSP with multiple popular generative ZSL models on three benchmark datasets. Each row pair shows the effect of adding DSP to a particular generative ZSL model.}
	\vspace{2mm}
	\resizebox{0.9\textwidth}{!}
	{
		\begin{tabular}{l|ccl|ccl|ccl}
			\hline
			\multirow{2}{*}{Generative ZSL Methods} 
			&\multicolumn{3}{c|}{\textbf{CUB}}&\multicolumn{3}{c|}{\textbf{SUN}}&\multicolumn{3}{c}{\textbf{AWA2}}\\
			\cline{2-4}\cline{5-7}\cline{8-10}
			&\rm{U} & \rm{S} & \rm{H} &\rm{U} & \rm{S} & \rm{H} &\rm{U} & \rm{S} & \rm{H}  \\
			\hline
			CLSWGAN~\cite{Xian2018FeatureGN}&43.7&57.7&49.7&42.6&36.6&39.4&57.9&61.4&59.6\\
			CLSWGAN~\cite{Xian2018FeatureGN}+\textbf{DSP}&\textbf{51.4}&\textbf{63.8}&	\textbf{56.9}\textbf{$^{\color{blue}\uparrow\text{\textbf{7.2}}}$}& \textbf{48.3}&\textbf{43.0}&\textbf{45.5}$^{\color{blue}\uparrow\text{\textbf{6.1}}}$&\textbf{60.0}&\textbf{86.0}&\textbf{70.7}$^{\color{blue}\uparrow\text{\textbf{11.1}}}$\\
			\cline{1-10}
			f-VAEGAN~\cite{Xian2019FVAEGAND2AF}&48.7&58.0&52.9&45.1&38.0&41.3&57.6&70.6&63.5\\
			f-VAEGAN~\cite{Xian2019FVAEGAND2AF}+\textbf{DSP}&\textbf{62.5}&\textbf{73.1}&\textbf{67.4}\textbf{$^{\color{blue}\uparrow\text{\textbf{14.5}}}$}& \textbf{57.7}&\textbf{41.3}	&\textbf{48.1}$^{\color{blue}\uparrow\text{\textbf{6.8}}}$&\textbf{63.7}&\textbf{88.8}&\textbf{74.2}$^{\color{blue}\uparrow\text{\textbf{10.7}}}$\\
			\cline{1-10}
			TF-VAEGAN~\cite{Narayan2020LatentEF}&53.7&61.9&	57.5&48.5&37.2&42.1&58.7&76.1&66.3\\
			TF-VAEGAN~\cite{Narayan2020LatentEF}+\textbf{DSP}&\textbf{58.7}&\textbf{67.4}&\textbf{62.8}$^{\color{blue}\uparrow\text{\textbf{5.3}}}$&\textbf{60.3}&\textbf{45.3}&\textbf{51.7}$^{\color{blue}\uparrow\text{\textbf{9.6}}}$&\textbf{65.6}&\textbf{87.1}&\textbf{74.8}$^{\color{blue}\uparrow\text{\textbf{8.5}}}$\\
			\cline{1-10}
			FREE~\cite{Chen2021FREE}&54.9&60.8&57.7&47.4&37.2&41.7&60.4&75.4&67.1\\
			FREE~\cite{Chen2021FREE}+\textbf{DSP}&\textbf{60.9}&\textbf{68.7}&\textbf{64.6}$^{\color{blue}\uparrow\text{\textbf{6.9}}}$&\textbf{60.3}&\textbf{44.1}&\textbf{51.0}$^{\color{blue}\uparrow\text{\textbf{9.3}}}$&\textbf{65.3}&\textbf{89.2}&\textbf{75.4}$^{\color{blue}\uparrow\text{\textbf{8.3}}}$
			\\
			\hline
	\end{tabular} }
	\label{table:baseline+dsp} 
\end{table*}

\subsection{Ablation Study}
We train our DSP via loss terms $\mathcal{L}_{Scyc}$, $\mathcal{L}_{V2S}$, and $\mathcal{L}_{S2S}$. The prototype evolves in a smooth manner, as shown in Eq.~\ref{eq:DSP}. Besides, feature enhancement via prototype concatenation is utilized for inference. We validate the effectiveness of loss terms, smooth evolvement, and feature enhancement by using the f-VAEGAN as the baseline. Table \ref{table:ablation} shows the ablation results. When DSP is without $\mathcal{L}_{Scyc}$ during training, the performance degrades as shown in the second row (i.e., w/o $\mathcal{L}_{Scyc}$). This is because the generator lacks supervision from semantic cycle consistency. On the other hand, the results do not drop significantly without $\mathcal{L}_{S2S}$. However, if DSP does not contain $\mathcal{L}_{V2S}$, the performance drops significantly (i.e., the harmonic mean drops by 7.5\% and 5.4\% on CUB and SUN, respectively). These results show that the supervision from V2SM outputs (i.e., prototypes mapped from sample features) is effective for prototype evolvement in VOPE. The prototypes are gradually aligned to the real prototypes by matching the visual sample features via loss terms, which mitigates the visual-semantic domain shift problem in generative ZSL. 

Besides loss terms, the dynamic prototype evolvement via smooth averaging is validated in our study. Without smooth evolvement (i.e., moving average in Eq.~\ref{eq:DSP}), the performance of our DSP decreases significantly (i.e., the harmonic mean drops by 12.6\% and 6.0\% on CUB and SUN, respectively, compared to the full DSP on the last row). In the inference phase, feature enhancement via prototype concatenation improves the DSP performance. The results in Table \ref{table:ablation} show that our DSP method is effective in mitigating the visual-semantic domain shift problem and achieves favorable performance.

{\flushleft \bf Generative ZSL methods with DSP.}
Our DSP is a general prototype evolvement method for generative ZSL. We experimentally analyze whether our DSP improves existing generative ZSL methods. We introduce four prevalent generative ZSL methods, including CLSWGAN~\cite{Xian2018FeatureGN}, f-VAEGAN~\cite{Xian2019FVAEGAND2AF}, TF-VAEGAN~\cite{Narayan2020LatentEF}, and FREE~\cite{Chen2021FREE}. Their official implementations are utilized for producing results. We note that TF-VAEGAN and FREE contain the semantic decoder. So we only integrate VOPE with $\mathcal{L}_{V2S}$ and $\mathcal{L}_{S2S}$ for these two generative ZSL methods during the training phase. The detailed settings of our DSP integrated on these generative ZSL methods are in Sec.~\ref{appdx-C}. Table \ref{table:baseline+dsp} shows the evaluation results. Our DSP consistently improves these generative ZSL methods on all benchmark datasets by a large margin. The average performance gains of harmonic mean are 8.5\%, 8.0\%, and 9.7\% on CUB, SUN, and AWA2, respectively. These results show that visual-semantic domain shift is an important but unnoticed problem in existing generative ZSL methods. Our DSP evolving method is effective in improving the performance of existing generative ZSL methods.

\begin{figure*}[t]
	\begin{center}
		\begin{tabular}{cc}
			\includegraphics[width=0.4\linewidth]{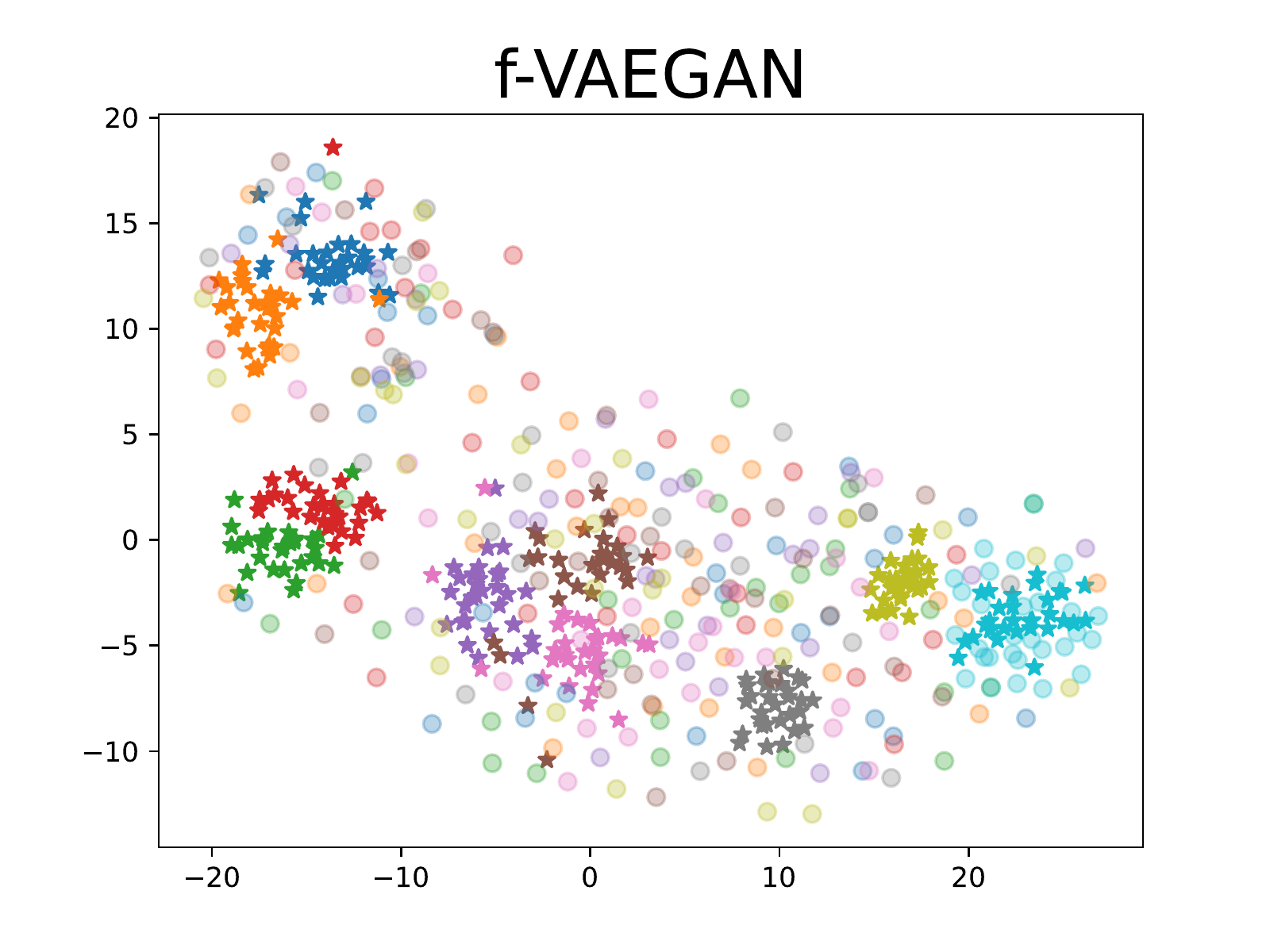}&
			\includegraphics[width=0.4\linewidth]{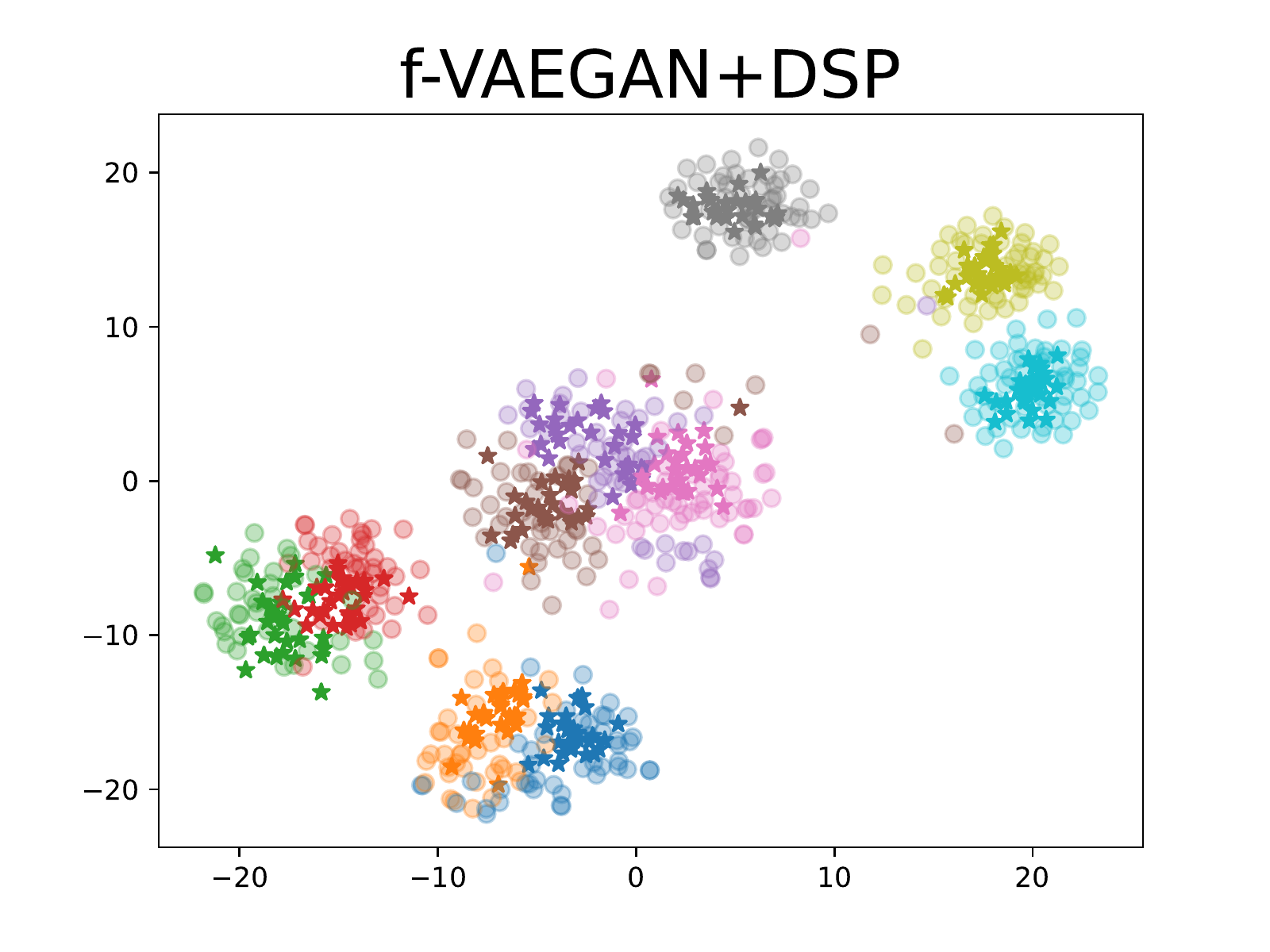}\\ 
		\end{tabular}
		\vspace{-4mm}
		\caption{Qualitative evaluation with t-SNE visualization. The sample features from f-VAEGAN are shown on the left, and from f-VAEGAN with our DSP integration are shown on the right. We use 10 colors to denote randomly selected 10 classes from CUB. The $\circ$ and $\star$ are denoted as the real and synthesized sample features, respectively. The synthesized sample features and the real features distribute differently on the left while distributing similarly on the right. (Best Viewed in Color)}\vspace{-3mm}
		\label{fig:t-SNE}
	\end{center}
\end{figure*}

\renewcommand{\tabcolsep}{1pt}
\begin{figure*}
	\begin{center}
		\begin{tabular}{cccc}
			\vspace{-2mm}
			\includegraphics[width=0.25\linewidth]{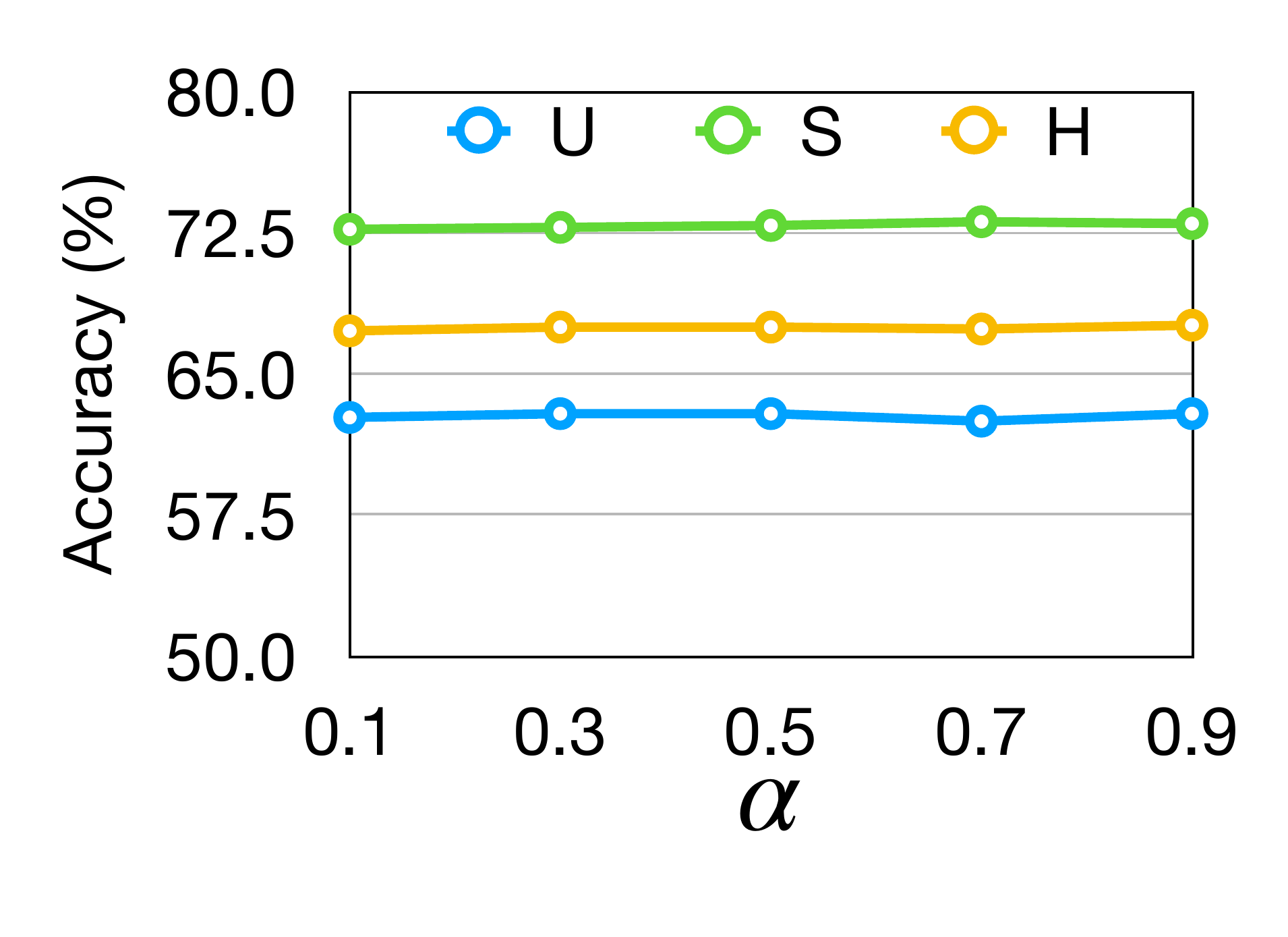}&
			\includegraphics[width=0.25\linewidth]{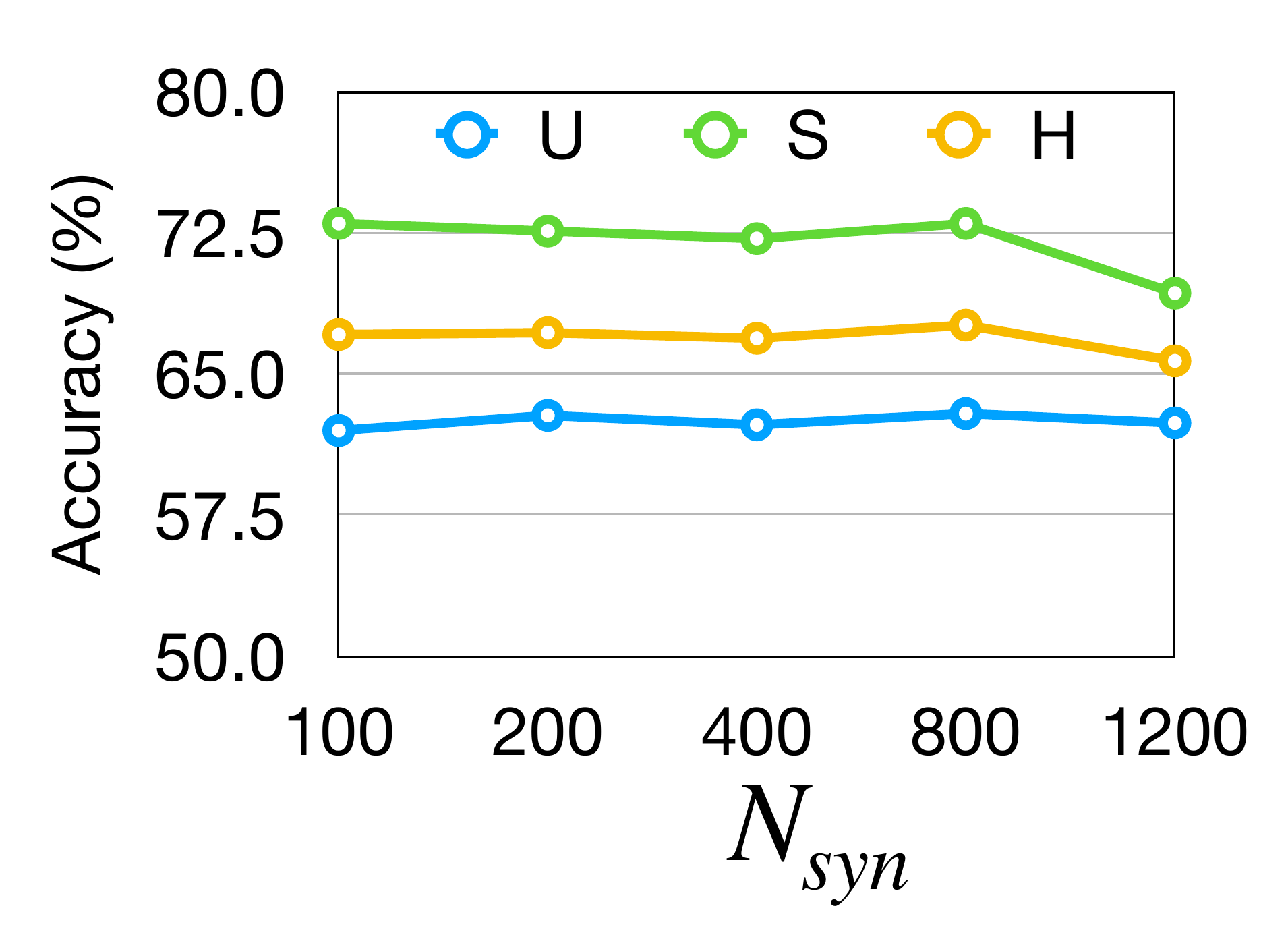}&
			\includegraphics[width=0.25\linewidth]{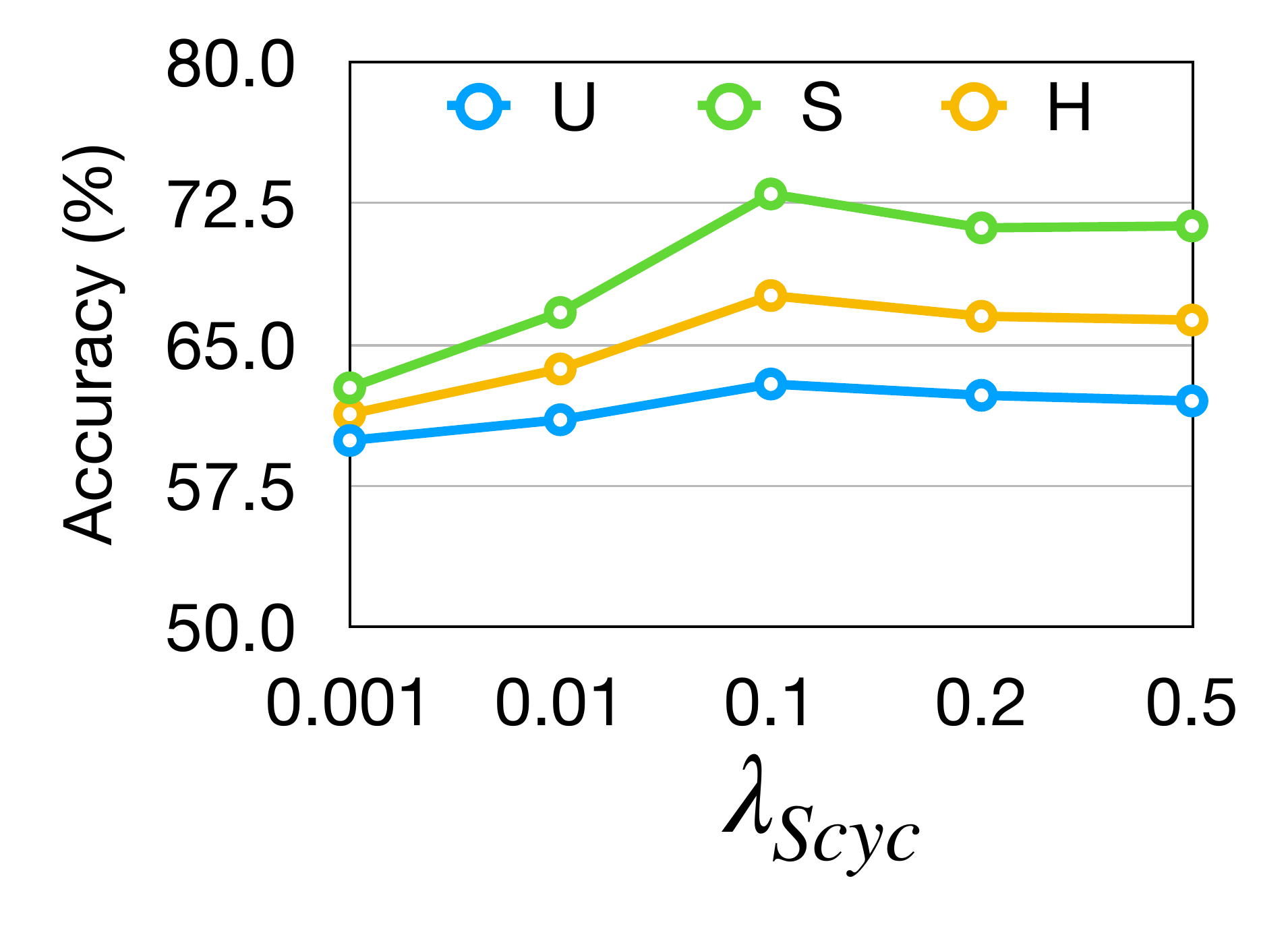}&
			\includegraphics[width=0.25\linewidth]{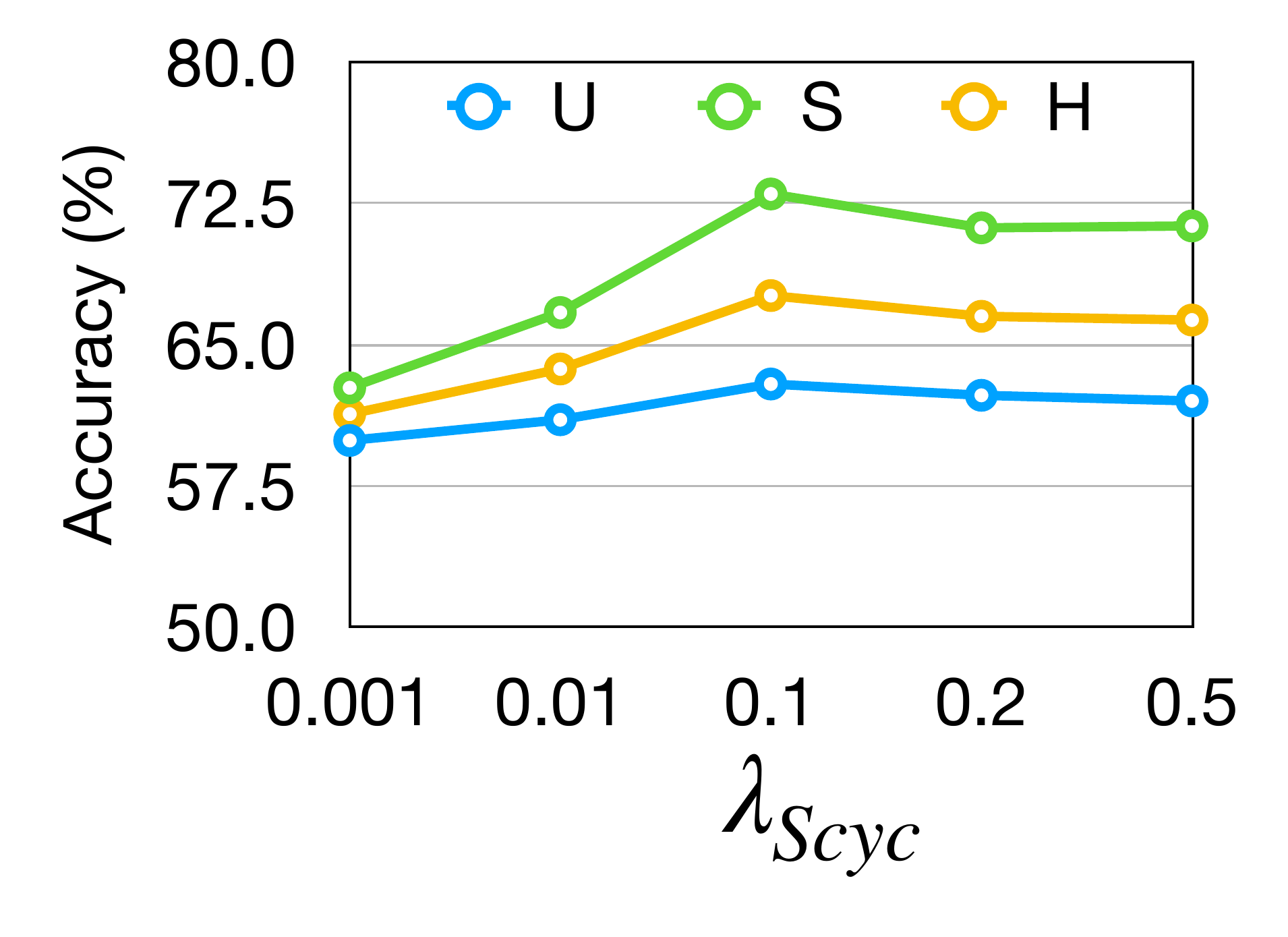}\\
			(a) Varying $\alpha$ effect& (b) Varying $N_{syn}$ effect& 
			(c) Varying $\lambda_{Scyc}$ effect& (d) Varying $\lambda_{V2S}$ effect\\
		\end{tabular}
		\vspace{-2mm}
		\caption{Hyper-parameter analysis. We show the GZSL performance variations on CUB by adjusting the value of $\alpha$ in (a), the value of $N_{syn}$ in (b), the value of loss weight $\lambda_{Scyc}$ in (c), and the value of loss weight $\lambda_{V2S}$ in (d). (Best Viewed in Color)}
		\label{fig:hyper-para}
	\end{center}
\end{figure*}

{\flushleft \bf Qualitative evaluation.}
We conduct a qualitative evaluation of the performance improvement of our DSP integrated into the baseline f-VAEGAN. The t-SNE visualization~\cite{Maaten2008VisualizingDU} of the real and synthesized sample features is shown in Fig. \ref{fig:t-SNE}. We randomly select 10 classes from CUB, and visualize these two types of sample features of f-VAEGAN on the left. Then, we visualize the same features of f-VAEGAN+DSP on the right. The left figure shows that sample features synthesized by f-VAEGAN and the real features distribute very differently, which brings the visual-semantic domain shift problem. In contrast, our DSP evolves the predefined semantic prototypes to match the real sample features. The prototype evolvement enables the generator to synthesize sample features close to the real features, as shown on the right figure. The visualization shows that our DSP helps the generative ZSL method to synthesize sample features close to the real sample features, which benefits classifier training and improves ZSL performance.

{\flushleft \bf Hyper-parameter analysis.}		
We analyze the effects of different hyper-parameters of our DSP under the CUB dataset. These hyper-parameters include the smooth fusion weight $\alpha$ in Eq.~\ref{eq:DSP}, the number of synthesized samples for each unseen class $N_{syn}$, and the loss weight $\lambda_{Scyc}$ and $\lambda_{V2S}$. We set $\lambda_{Scyc}=\lambda_{S2S}$ as both of them are the semantic reconstruction loss, and we take one loss (i.e., $\lambda_{Scyc}$) for analysis here. Fig.~\ref{fig:hyper-para} shows the ZSL performance of using different hyper-parameters. In (a), the results show that DSP is robust to varying values of $\alpha$ and achieves good performance when $\alpha$ is relatively large (i.e., $\alpha=0.9$). This is because of the smooth prototype evolvement that does not change significantly for each step. In (b), our DSP is shown robust to $N_{syn}$ when it is not set in a large number. The $N_{syn}$ can be set as 800 to balance between the data amount and the ZSL performance. In (c) and (d), we conclude that we set $\lambda_{Scyc}$ as 0.1, and set $\lambda_{V2S}$ as 0.6 to achieve good performance. Moreover, we find that the value of $\lambda_{V2S}$ should be larger than those of $\lambda_{Scyc}$ and $\lambda_{S2S}$. This indicates that using V2SM output (i.e., prototypes mapped via semantic sample features) to supervise VOPE output is the crucial design to match the evolved prototypes and sample features. Under this supervision, the evolvement of semantic prototypes gradually matches the sample features synthesized by the generator $G$. Overall, Fig.~\ref{fig:hyper-para} shows that our DSP is robust to overcome hyper-parameter variations. 

\section{Concluding Remarks}\label{Sec5}
In this work, we analyze that the predefined semantic prototype introduces the visual-semantic domain shift problem in generative ZSL. We then propose a dynamic semantic prototype evolving method to mutually refine the prototypes and sample features. After evolvement, the generator synthesizes sample features close to the real ones and benefits ZSL classifier training. Our DSP can be integrated into a series of generative ZSL methods for performance improvement. Experiments on the benchmark datasets indicate our effectiveness, which shows that evolving semantic prototype explores one promising virgin field in generative ZSL.

\section*{Acknowledgements}
The work was supported in part by the NSF-Convergence Accelerator Track-D award \#2134901, by the NSFC~(62172177), by the National Key R\&D Program (2022YFC3301004,2022YFC3301704), by the National Institutes of Health (NIH) under Contract R01HL159805, by grants from Apple Inc., KDDI Research, Quris AI, and IBT, and by generous gifts from Amazon, Microsoft Research, and Salesforce.

%\noindent \textbf{Limitations}: We find that our DSP indirectly conveys the visual information into VOPE, resulting in the visual information cannot significantly guide the evolution of the predefined semantic prototypes. We could design a directed way to convey the visual signal for supervising the evolution of semantic prototypes in the future.

%\clearpage
\bibliography{mybibfile}
\bibliographystyle{icml2023}

%%%%%%%%%%%%%%%%%%%%%%%%%%%%%%%%%%%%%%%%%%%%%%%%%%%%%%%%%%%%%%%%%%%%%%%%%%%%%%%
%%%%%%%%%%%%%%%%%%%%%%%%%%%%%%%%%%%%%%%%%%%%%%%%%%%%%%%%%%%%%%%%%%%%%%%%%%%%%%%
% APPENDIX
%%%%%%%%%%%%%%%%%%%%%%%%%%%%%%%%%%%%%%%%%%%%%%%%%%%%%%%%%%%%%%%%%%%%%%%%%%%%%%%
%%%%%%%%%%%%%%%%%%%%%%%%%%%%%%%%%%%%%%%%%%%%%%%%%%%%%%%%%%%%%%%%%%%%%%%%%%%%%%%
\newpage
\appendix
\onecolumn

{\centering \textbf{APPENDIX}}\\
Appendix organization:
\begin{itemize}
	\item {\color{red}Appendix} \ref{appdx-A}: Network details of V2SM.
	\item {\color{red}Appendix} \ref{appdx-B}: Network details of VOPE.
	\item {\color{red}Appendix} \ref{appdx-C}: The hyper-parameter settings of our DSP entailed on four baselines.
	%\item {\color{red}Appendix} \ref{appdx-D}:  t-SNE Visualization on SUN and AWA2.
\end{itemize}
\section{Nework Details of V2SM}\label{appdx-A}
\begin{figure}[h]
	\begin{center}
		\vspace{-1mm}
		\includegraphics[width=1.0\linewidth]{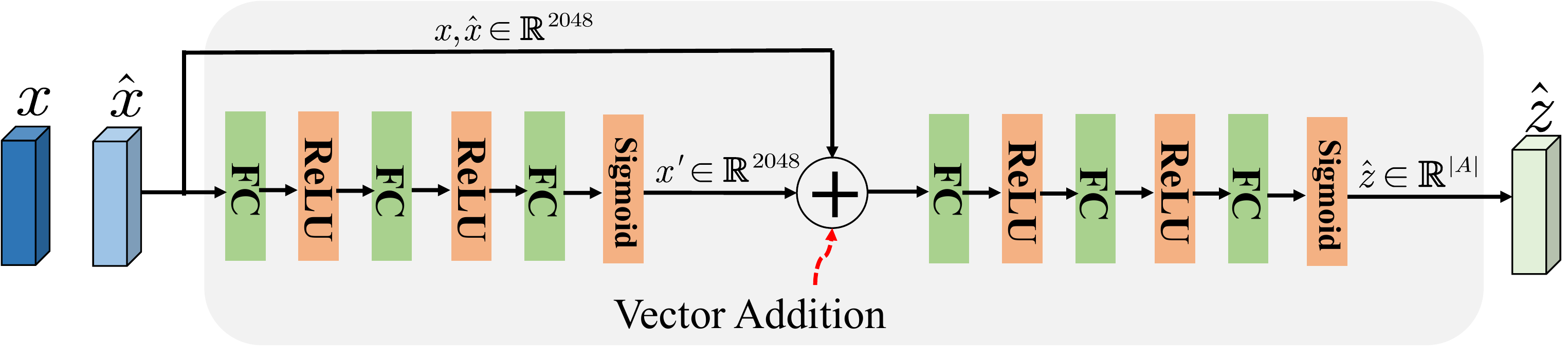}\\
		\vspace{-2mm}
		\caption{Network details of V2SM.}
		\label{fig:V2SM}
	\end{center}\vspace{-4mm}
\end{figure}

As shown in Fig. \ref{fig:V2SM}, the network of V2SM is an MLP with a residual block, which avoids too much information loss from 2048-dim to $|A|$-dim.
\section{Nework Details of VOPE}\label{appdx-B}
\begin{figure}[h]
	\begin{center}
		\vspace{-1mm}
		\includegraphics[width=0.65\linewidth]{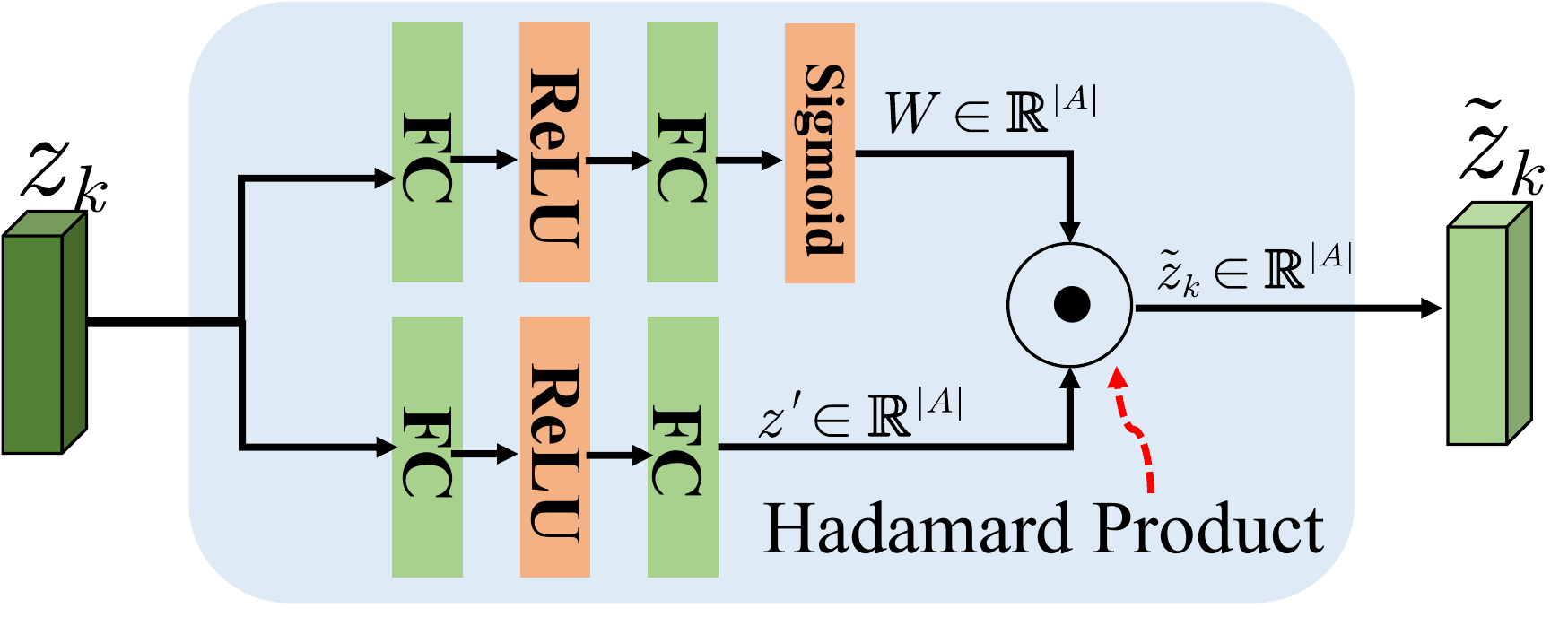}\\
		\vspace{-2mm}
		\caption{Network details of VOPE.}\vspace{-4mm}
		\label{fig:VOPE}
	\end{center}
\end{figure}

As shown in Fig. \ref{fig:VOPE}, the network of VOPE is an MLP with a residual block (fusing with Hardamad Product), which preserves enough semantic information of the predefined semantic prototype and enables VOPE to evolve progressively.

\section{The Hyper-Parameter Settings of Our DSP Entailed on Various Baselines}\label{appdx-C}

We present the hyper-parameter settings of our DSP entailed on various baselines (\textit{i.e.},  CLSWGAN~\cite{Xian2018FeatureGN}, f-VAEGAN~\cite{Xian2019FVAEGAND2AF}, TF-VAEGAN~\cite{Narayan2020LatentEF} and FREE~\cite{Chen2021FREE}) on CUB, SUN and AWA2. Including the synthesizing number of per unseen classes $N_{syn}$, the loss weights $\lambda_{Scyc}$, $\lambda_{V2S}$ and combination coefficient $\alpha$ in Eq. \ref{eq:DSP}.  We empirically observe that our DSP is robust and easy to train when it is entailed on various generative models. Based on these hyper-parameter settings, our DSP achieves significant performance gains over the various popular generative models on all datasets. For instance, the average performance gains of harmonic mean are 8.5\%, 8.0\% and 9.7\% on CUB, SUN and AWA2, respectively. Please refer to Table \ref{fig:hyper-para}.
\begin{table*}[ht]
	\small
	\centering  
	\caption{The hyper-parameter settings of our DSP entailed on various baselines (\textit{i.e.},  CLSWGAN~\cite{Xian2018FeatureGN}, f-VAEGAN~\cite{Xian2019FVAEGAND2AF}, TF-VAEGAN~\cite{Narayan2020LatentEF} and FREE~\cite{Chen2021FREE}) on CUB, SUN and AWA2. Including the synthesizing number of per unseen classes $N_{syn}$, the loss weights $\lambda_{Scyc}$, $\lambda_{V2S}$ and combination coefficient $\alpha$  in  Eq. \ref{eq:DSP}.}
	\resizebox{\linewidth}{!}{\small
		\begin{tabular}{l|cccc|cccc|cccc}
			\hline
			\multirow{2}{*}{\textbf{Methods}} 
			&\multicolumn{4}{c|}{\textbf{CUB}}&\multicolumn{4}{c|}{\textbf{SUN}}&\multicolumn{4}{c}{\textbf{AWA2}}\\
			\cline{2-5}\cline{6-9}\cline{9-13}
			&\rm{$N_{syn}$}&\rm{$\lambda_{Scyc}$} & \rm{$\lambda_{V2S}$} & \rm{$\alpha$}&\rm{$N_{syn}$}&\rm{$\lambda_{Scyc}$} & \rm{$\lambda_{V2S}$} & \rm{$\alpha$} &\rm{$N_{syn}$}&\rm{$\lambda_{Scyc}$} & \rm{$\lambda_{V2S}$} & \rm{$\alpha$} \\
			\hline
			clswGAN~\cite{Xian2018FeatureGN} + \textbf{DSP}	&300&0.15&1.0&0.9&300&0.005&1.0&0.9&3400&0.1&1.0&0.9\\
			\hline
			f-VAEGAN~\cite{Xian2019FVAEGAND2AF}+\textbf{DSP}& 800&	0.1&0.6&0.9&150&0.01&1.0&0.9&3400&0.001&0.6&0.9\\
			\hline
			TF-VAEGAN~\cite{Narayan2020LatentEF}+ \textbf{DSP}&400&0.01&1.0&0.9&500&0.05&1.5&0.9&5300&0.09&1.4&0.9\\
			\hline
			FREE~\cite{Chen2021FREE} + \textbf{DSP}&600&0.1&0.6&0.9&150&0.01&1.0&0.9&4000&0.001&2.0&0.9\\
			\hline
	\end{tabular} }
	\label{table:hyper-para}
\end{table*}

\end{document}